\pdfoutput=1

\documentclass[11pt]{article}

\usepackage{acl2024}
\usepackage{multirow}
\usepackage{amssymb}
\usepackage{ragged2e}
\usepackage{times}
\usepackage{latexsym}
\usepackage{graphicx}
\usepackage{tikz}
\usepackage{float}
\usepackage{pgfplots}
\pgfplotsset{compat=1.18, width=7.7cm}
\usepackage[T1]{fontenc}
\usepackage{enumitem}
\usepackage{makecell}

\usepackage[utf8]{inputenc}

\usepackage{subfig}
\usepackage{overpic}   
\usepackage{float}

\usepackage{microtype}

\usepackage{inconsolata}
\usepackage{booktabs}

\usepackage[many,most]{tcolorbox}

\newtcolorbox{mybox}{%
  breakable,enhanced,colback=white,colframe=black,left=0.5em,right=0.5em,boxrule=1.0pt}

\newtcolorbox{mybox3}[1]{%
  breakable,enhanced,colbacktitle=white,coltitle=black,colback=white,colframe=black,fonttitle=\bfseries,title=#1,leftupper=0.5em,rightupper=0.5em,boxrule=1.0pt}

\usepackage[first=0,last=9]{lcg}
\usepackage{colortbl}
\definecolor{Gray}{gray}{0.8}
\colorlet{Red}{red!10!white}
\colorlet{Blue}{blue!10!white}

\title{ERA-CoT: Improving Chain-of-Thought through Entity Relationship Analysis}

\author{
    ~~Yanming Liu$^{1}$
   ~~Xinyue Peng$^{2}$
   ~~Tianyu Du$^{1}$\footnotemark[1]
   ~~Jianwei Yin$^{1}$
   ~~Weihao Liu 
   ~~Xuhong Zhang$^{1}$\\ 
   $^{1}$Zhejiang University \\
   $^{2}$Southeast University \\
   \texttt{{\{oceann24, zhangxuhong, zjradty\}@zju.edu.cn, zjuyjw@cs.zju.edu.cn,}}\\
\texttt{{xinyuepeng@seu.edu.cn, liuweihao2022@outlook.com}}\\
}

\begin{document}
\maketitle

\renewcommand{\thefootnote}{\fnsymbol{footnote}}
\footnotetext[1]{Corresponding author.}
\renewcommand{\thefootnote}{\arabic{footnote}}

\begin{abstract}
Large language models (LLMs) have achieved commendable accomplishments in various natural language processing tasks. 
However, LLMs still encounter significant challenges when dealing with complex scenarios involving multiple entities. 
These challenges arise from the presence of implicit relationships that demand multi-step reasoning. 
In this paper, we propose a novel approach ERA-CoT, which aids LLMs in understanding context by capturing relationships between entities and supports the reasoning of diverse tasks through Chain-of-Thoughts (CoT). 
Experimental results show that ERA-CoT demonstrates the superior performance of our proposed method compared to current CoT prompting methods, achieving a significant improvement of an average of 5.1\% on GPT3.5 compared to previous SOTA baselines. Our analysis indicates that ERA-CoT increases the LLM's understanding of entity relationships, significantly improves the accuracy of question answering, and enhances the reasoning ability of LLMs.\footnote{Our code is public at \url{https://github.com/OceannTwT/era-cot}}

\end{abstract}

\section{Introduction}

Large language models (LLMs)~\citep{hoffmann2022training,chowdhery2023palm,touvron2023llama} have shown remarkable in-context learning capabilities in various natural language processing (NLP) tasks, including machine translation~\citep{vilar2022prompting,moslem2023adaptive}, question answering~\citep{robinson2022leveraging,li2022self,lazaridou2022internet}, and named entity extraction~\citep{chowdhery2023palm,brown2020language}, etc. 
Recently, prompting strategies like Chain-of-Thought (CoT)~\citep{wei2022chain} have garnered attention due to their capacity to significantly enhance LLMs reasoning capabilities. 
Considering the ability of CoT to guide LLMs in breaking down complex reasoning processes into simple steps, it stands out compared to standard zero-shot and few-shot methods. 

However, due to the presence of numerous entities such as characters, locations, etc., and the multitude of implicit relationships among them in certain scenarios, CoT still faces significant challenges in handling these situations. Named Entity Recognition (NER) has typically been employed when addressing these tasks. NER is a sequence labeling task in nature, where the model needs to assign an entity-type label to each token within a sentence \citep{wang2023gpt}.
Relation extraction is a category of methods for handling entity relationships within text passages. Various studies \citep{li2023revisiting, zhang-etal-2023-aligning} have also investigated the performance of LLMs in zero-shot relation extraction. However, without additional prompts,  LLMs have limited entity and relation extraction capabilities. Considering the importance of contextual content in answering questions, addressing knowledge-intensive tasks also requires a comprehensive analysis of entity relationships. 

In this paper, we propose Entity Relationship Analysis with Chain-of-Thought (ERA-CoT), a novel framework to better address reasoning tasks in complex entity scenarios. First, we extract all the entities involved in the text; second, we extract the directly mentioned explicit relationships between entities based on the text; then, we infer the indirect implicit relationships between entities based on these explicit relationships and the hidden information in the text; after that, we let the model score the implicit relationships based on the reliability of the relationships, set a threshold for judging the reliability of the relationships, and eliminate the implicit relationships that are lower than the threshold; finally, answer the questions based on the previously extracted entities and the obtained implicit and explicit relationships.

We conducted experiments on six widely adopted datasets and compared with four baseline methods.
The results show that ERA-CoT outperforms baselines on nearly all benchmarks, achieving a significant improvement of about 5.1\% on average. From the performance results, our method outperforms on all three types of reasoning problems: commonsense reasoning, mathematical reasoning, and logical reasoning. This indicates that enhancing the model's understanding of entity relationships can significantly boost the reasoning abilities and accuracy in answering questions of LLMs.
Our main contributions can be summarized as follows.
\begin{itemize}[leftmargin=1em]
\item[$\bullet$] We introduce ERA-CoT, a novel framework designed to conduct relationship analysis among multiple entities within complex scenarios during the zero-shot problem-solving process, which significantly strengthens the reasoning and comprehension abilities of LLMs.
\item[$\bullet$] 
Our method extends entity relationship analysis and relation extraction to CoT. It is capable of both further complex relationship inference after entity extraction in NER, and step-by-step accurate logical analysis for any complex scenario on a zero-shot setting.
\item[$\bullet$] Compared to baselines, we achieved an accuracy improvement of approximately 7.3\%. Our approach excels not only on GPT-3.5 but also demonstrates significant improvements on the open-source large model Llama-2. This indicates the versatility of our method for problem reasoning across various models and scenarios.

\end{itemize}
\section{Related Work}
\subsection{Chain of thought}
To utilize LLMs to solve more complex and logical reasoning tasks, \citet{wei2022chain} extended in-context learning by introducing the concept of Chain of Thought (CoT) through a step-by-step reasoning process. \citet{kojima2022large} found that simply adding a leading sentence ``Let's think step by step'' to a cue allowed LLMs to perform zero-shot logical reasoning without any additional human prompts \citep{chu2023survey}. Subsequently, CoT-SC \citep{wang2023selfconsistency} introduces a self-consistency strategy to replace the greedy decoding strategy. Auto-CoT\citep{zhang2023automatic} automatically constructs CoT based on questions, eliminating the instability of manual prompts. Complex-CoT\citep{fu2023complexitybased} employs multi-step reasoning estimation on CoT based on complexity. RE2\citep{xu2023rereading} utilizes a question rephrasing strategy to enhance the model's understanding of questions with zero prompts. \citet{Wang2023PlanandSolvePI} breaks down the problem into planning and solving steps to generate and answer the Chain-of-Thought. These studies highlight the importance of CoT in enhancing LLMs' reasoning and planning abilities in complex situations. However, further refinement of CoT is needed in scenarios involving complex relationships with multiple entities.

\subsection{Named Entity Recognition}
Named Entity Recognition (NER) is the task of identifying mentions of entities from unstructured text and categorizing them properly \citep{moscato2023few}. NER not only serves as a standalone tool for Information Extraction (IE) but also plays a crucial role in various NLP applications, such as text understanding \citep{zhangernie,cheng2020attending}, information retrieval \citep{luo2020hierarchical,taille2020contextualized}, question answering \citep{luo2020bert}, machine translation \citep{malmasi2022multiconer}, knowledge base construction \citep{tabassum2020code}, etc.

Recently, with the popularity of LLMs, NER has seen more profound development. \citet{malmasi2022multiconer} presents a large multilingual dataset to represent contemporary challenges including low-context scenarios, and syntactically complex entities in NER. \citet{wang2023gpt} proposes GPT-NER to resolve the gap by transforming the sequence labeling task to a generation task that can be easily adapted by LLMs. \citet{das2022container} and \citet{ashok2023promptner} study few-shot NER and demonstrate the great performance of few-shot NER in prompt engineering and cross-domain NER. \citet{li2023leveraging} proposed a graph based on entities and relationships, and answered questions based on graph dependencies. This paper will go further by analyzing the relationships between entities.

\begin{figure*}[tp]
    \centering
    \includegraphics[width=0.98\linewidth]{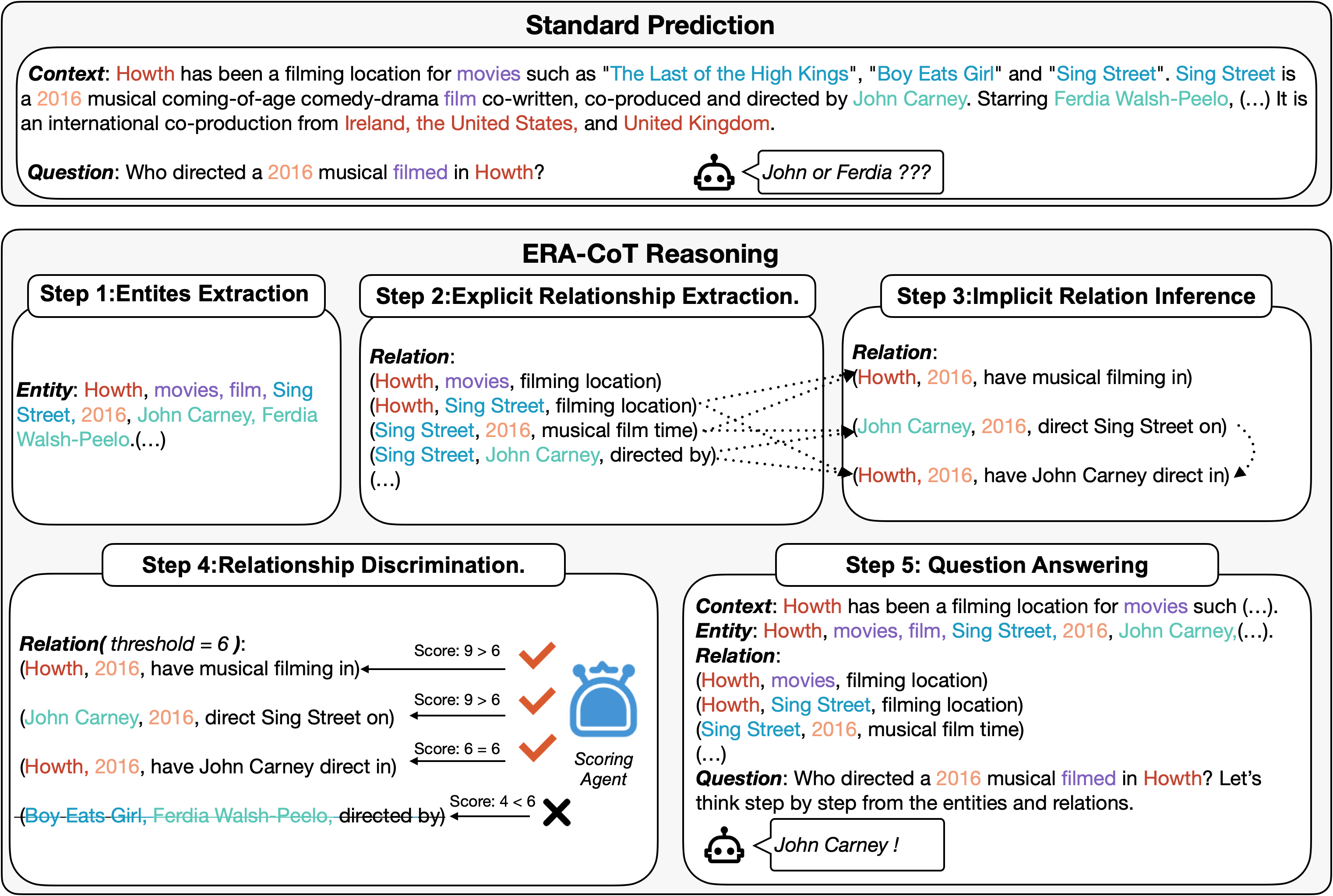}
    \caption{The top of the figure represents the standard prediction process. The bottom of the figure shows the five-step inference process of ERA-CoT, which relies on the extraction of entities and the inference and analysis of relationships between entities to obtain the results.}
    \vspace{-0.15cm}
    \label{fig:enter-label}
\end{figure*}

\subsection{Relation Extraction}

Relation extraction involves extracting relationships based on relevant context, extracting the relationship between two given entities. It plays a crucial role in information extraction and knowledge construction.
Traditional relation extraction involves fine-tuning pre-trained language models to learn relationship features, thereby providing solutions \citep{han-etal-2019-opennre, zhou-chen-2022-improved, lyu-chen-2021-relation, xu-etal-2023-unleash} for relation extraction in the few-shot setting.
Some studies \citep{jimenez-gutierrez-etal-2022-thinking, wei2023zeroshot} have explored the performance of relation extraction on large-scale models and discussed the drawbacks of large models in relation extraction. \citet{zhang-etal-2023-aligning} aligns relation extraction with question answering, obtaining corresponding relationships through question and answer based on relation templates. \citet{li2023revisiting} transforms relation extraction into a multi-stage question-answering process, offering a novel approach to address zero-shot relation extraction. \citet{wan2023gpt} give a solution to relation extraction under a few-shot prompting setting.
Our approach combines Chain-of-Thought with relation extraction by optimizing the relation extraction process. Through stage-wise relation extraction, it enhances the ability to extract relationships while strengthening problem-solving capabilities.

\section{Methodology}

As shown in Figure \ref{fig:enter-label}, we introduce ERA-CoT, a novel framework to enhance the model's understanding and reasoning of entity relationships in various NLP tasks. It consists of five stages, each involving different degrees of enhanced understanding of entity relations. Simultaneously, leveraging the model's capabilities, it can jointly learn explicit and implicit relationships in the context throughout this progressive process, filtering out relevant knowledge to enhance the capability and performance of question reasoning. 

\vspace{5pt}

{
\setlength{\parindent}{0cm}
\textbf{Problem Formulation.} Given an input sequence $x$, the CoT method is to predict the answer via:
\begin{equation} 
y = {\arg\max\limits_{y_i}} P(y_i|\mathcal{T}, x),
\end{equation}
where $\mathcal{T}$  is the task instruction and $y_i$ indicate all possible results of $y$.
Our framework is to optimize the process based on the following steps.
}

\vspace{5pt}

{
\setlength{\parindent}{0cm}

\textbf{Step 1: Entities Extraction.} 
Leveraging the information extraction capability of LLMs \citep{ashok2023promptner}, we present the sentence to the model and request it to extract all $n$ entities $ \mathcal{E} = \{(s_i,t_i)\}^n_{i=1}$ from the provided text, expressing them as a pair relationship $e_i = (s_i,t_i)$. These relationships encompass specific information about the entities span $s_i$ and their corresponding entity type $t_i$. Formally, we provide an input sentence to the large model, utilizing its NER capability to predict the corresponding entity spans and classifications. The option of entity types is related to a predefined entity set $S$.

}

Additionally, we adopt Self-Consistency (SC) \citep{narangself} to evaluate the consistency of NER, as an entity is extracted from the query $x$, it is measured by LLM to verify $n$ times to determine whether it is an entity. If the upvote is higher than $\left\lceil{\frac{n}{2}}\right\rceil$, the entity is deemed as a valid entity. Such a design could help to remove false NER extraction.

\vspace{5pt}

{
\setlength{\parindent}{0cm}

\textbf{Step 2: Explicit Relationship Extraction.} We aim to explore the relevant relationships between different entities in the zero-shot setting. While entities may have direct relationships in a sentence, we generate multiple pairs of relationships for each entity. Specifically, leveraging the contextual capabilities of the LLM, we extract all open relation entity pairs that are directly stated in the context. The relation could be simplified as triplets $(e_i, e_j, r)$ where entity $e_i$ and entity $e_j$ are sampled from texts and $r$ is the relations of these two entities. SC method is utilized to evaluate the consistency of explicit relation as previous step. The explicit relation set $\mathcal{R}_e$ could formulated as:
}
\begin{equation}
    \mathcal{R}_e = \bigcup\limits_{i,j \in \mathcal{E}}\{(e_i,e_j,r)\}.
\end{equation}
Through this process, the explicit relations are deconstructed into triplets formulation. As explicit relationships are relatively clear in the text, we use LLMs to extract explicit relationships from the text. Then, we can use these relations to help the following steps to find implicit entity relations. 

\vspace{5pt}
{
\setlength{\parindent}{0cm}
\textbf{Step 3: Implicit Relationships Inference.} We aim to perform entity relationship inference based on the preceding steps. While implicit relationships require multi-step inference, they are more challenging to discover compared to explicit relationships that can be directly extracted from the context. Therefore, we need to infer implicit relationships based on previously found explicit relationships. Specifically, we provide the original context $x$ along with all generated relationships $\mathcal{R}_0$ in the preceding steps, request LLMs to make reasoning, and generate $k$ most relevant relation. Assume that the intermediate relations are $T_i = (e_{_i}, e_{i+1}, r_i)$. It can help to generate reasonable relation triplets $T_{1 \rightarrow n} = ({e_1, e_n, r_k})$ from a relation chain $T_1$, $T_2$, ..., $T_{n-1}$, where $n$ is length of chain.
}

The procedure can be formalized using the following equation while $r_k$ indicates the $k$ different relations from LLMs reasoning:
\begin{equation}
    \mathcal{R}_i' = \bigcup\limits_{i,j \in \mathcal{E}}\{(e_{i},e_{j},r_k)\}.
\end{equation}

{
\setlength{\parindent}{0cm}

\textbf{Step 4: Relationship Discrimination.} Relying on the Self-Correction \citep{ganguli2302capacity} capability of the LLM, we set an LLM as a scoring agent. We provide the origin context and all triplets relations generated in Step 3 to the agent. Then, each relation gets a score from the agent. For triplets that are more likely to express the correct relationship, we assign a higher score to this relationship. A score threshold $v_{th}$ is established to assess whether the model correctly infers a relationship. Formally, for each triplet, the scoring agent model gives a value $\mathcal{V}(i,j,k)$ to assess the confidence level of the relation triplet $(e_{i}, e_{j}, r_k)$. We present more detailed scoring criterion on Appendix~\ref{sec:criterion}.
}

For scores below our pre-defined threshold $v_{th}$, we consider the relationships below the threshold scores as irrelevant. Relationships deemed irrelevant or incorrect are eliminated, and those remaining with higher scores are considered correct in the relationship discrimination process. This approach helps eliminate some erroneous relationships caused by model reasoning during relationship discrimination, thereby improving its consistency and accuracy. In this case, the implicit relation set could be stated as:
\begin{equation}
    \mathcal{R}_i = \bigcup\limits_{i,j \in \mathcal{E}, \mathcal{V}(i,j,k). \ge v_{th}}\{(e_{i},e_{j},r_k)\}.
\end{equation}

{
\setlength{\parindent}{0cm}
\textbf{Step 5: Question Answering.} Building upon all the relationships described above, we formalize their expressions and incorporate them with the original context into prompts. We utilize all entities, all relation triplets, and the context to predict the questions in this step. The ERA-CoT method could finally formulate as:
}
\begin{equation}
{y = {\arg\max\limits_{y_i}} P(y_i|\mathcal{R}=[\mathcal{R}_e,\mathcal{R}_i],\mathcal{E},\mathcal{T},S,x).}
\end{equation}



\begin{table*}[]
	 \resizebox{\linewidth}{!}{		
\begin{tabular}{l|l|cccccc}
\toprule
\textbf{Model}             & \textbf{Methods}                   & \textbf{StrategyQA} & \textbf{CSQA} & \textbf{LogiQA} & \textbf{HotpotQA} & \textbf{2WikiMHQA} & \textbf{GSM8K} \\ \midrule
\multirow{9}{*}{GPT3.5}   & Vanilla LM      & 65.4                & 72.1          & 28.2            & 49.7              & 55.7               & 52.5           \\
                           & CoT           & 63.2                & 77.2          & 36.4            & 52.4              & 61.2               & 70.4           \\
                           & CoT-SC@5      & 65.1                & 78.3          & 38.2            & 52.8              & 65.6               & 74.8           \\
                           & Auto-CoT      & 64.6                & 77.6          & 38.6            & 53.1              & 64.3               & 77.1           \\
                           & Complex-CoT      & 64.2                &76.2          & 38.6            & 52.5              & 65.3               & \textbf{80.1}           \\
                           & PS            & 65.7                & 77.5          & 37.8            & 53.2              & 63.8               & 76.2           \\
                           & PS+            & 66.2                & 77.1          & 38.9            & 53.7              & 64.5               & 75.8           \\
                           & RE2            & 67.1                & 79.3          & 39.5            & 53.3              & 65.4               & 76.5           \\
                           & \cellcolor{Red}ERA-CoT & \cellcolor{Red}\textbf{71.4}                & \cellcolor{Red}\textbf{83.2}          & \cellcolor{Red}\textbf{45.2}            & \cellcolor{Red}\textbf{58.4}              & \cellcolor{Red}\textbf{70.2}               & \cellcolor{Red}79.5           \\ \midrule
\multirow{9}{*}{$\text{Llama2}_\text{13B}$} & Vanilla LM      & 57.2                & 58.3          & 24.5            & 34.2              & 28.2               & 17.8           \\
                           & CoT           & 55.1                & 64.2          & 30.2            & 37.1              & 32.4               & 18.9           \\
                           & CoT-SC@5      & 57.2                & 66.8          & 32.4            & 36.8              & 34.6               & 21.2           \\
                           & Auto-CoT      & 56.8                & 66.5          & 31.9            & 37.5              & 35.2               & 20.1           \\
                           & Complex-CoT      & 54.8                &65.2          & 32.1            & 37.1              & 35.1               & 23.8           \\
                           & PS            & 56.8                & 66.2          & 31.6            & 36.9              & 34.2               & 22.4           \\
                           & PS+            & 57.6                & 66.9          & 32.4            & 36.7              & 34.8               & 23.1           \\
                           & RE2            & 58.4                & 67.5          & 33.1            & 37.5              & 36.1               & 22.9           \\
                           & \cellcolor{Red}ERA-CoT & \cellcolor{Red}\textbf{61.5}                & \cellcolor{Red}\textbf{72.6}          & \cellcolor{Red}\textbf{35.5}            & \cellcolor{Red}\textbf{39.2}              & \cellcolor{Red}\textbf{38.9}               & \cellcolor{Red}\textbf{24.5}           \\ \bottomrule
\end{tabular}}
    \caption{Main experimental results. The best results are highlighted in bold. We use accuracy as the evaluation metric. CoT-SC@5 represents retrieving five CoT reasoning chains to make majority votes.}
\label{tab:main_result}
\end{table*}

\section{Experimental Setup}

\subsection{Datasets and Models}
 We consider three reasoning scenarios, i.e., commonsense reasoning, logical reasoning, and mathematical reasoning.
Specifically, for commonsense reasoning, we use StrategyQA~\citep{geva2021did} and CSQA~\citep{talmor2019commonsenseqa}; for logical reasoning, we use LogiQA~\citep{liu2021logiqa}, HotpotQA~\citep{yang2018hotpotqa}, and 2WikiMultiHopQA~\citep{ho2020constructing}; for mathematical reasoning, we use GSM8K~\citep{cobbe2021training}.
For models, we use GPT3.5 (with 175 billion parameters)~\citep{openai2023gpt4} and Llama2$_\text{13B}$~\citep{touvron2023llama}.




\subsection{Baselines}

To evaluate our method holistically, we compare ERA-CoT with the leading CoT methods baselines:

\vspace{2pt}

{
\setlength{\parindent}{0cm} \textbf{Vanilla LM}, directly presents tasks and the corresponding questions, predicting the outcomes of the questions through in-context learning.
}

\vspace{5pt}

{
\setlength{\parindent}{0cm} \textbf{Chain-of-Thought} (CoT) \citep{wei2022chain}, predicts the answers by generating explanations and steps. 
}

\vspace{5pt}

{
\setlength{\parindent}{0cm} \textbf{CoT-SC} \citep{wang2023selfconsistency}, generates multiple paths of Chain-of-Thought and votes to select the highest-voted result as the final result. 
}

\vspace{5pt}

{
\setlength{\parindent}{0cm} \textbf{Auto-CoT} \citep{zhang2023automatic}, is a baseline that automatically generates multi-step reasoning in natural language.
}

\vspace{5pt}

{
\setlength{\parindent}{0cm} \textbf{Complex-CoT} \citep{fu2023complexitybased} utilizes a complexity-based strategy, sampling multiple chain-of-thoughts, and selecting answers that consistently align across complex inference chains through a majority vote.
}

\vspace{5pt}
{
\setlength{\parindent}{0cm} \textbf{PS and PS+} \citep{Wang2023PlanandSolvePI}, is a zero-shot CoT that breaks down the problem into planning and solving steps to generate the answers of Chain-of-Thought. PS+ extracts more details information like variables to help the inference process.
}

\vspace{5pt}

{
\setlength{\parindent}{0cm} \textbf{RE2} \citep{xu2023rereading}, a plug-and-play approach that entails re-reading the question before engaging in the reasoning process.
}









\subsection{Implementation}
We access the GPT models through the OpenAI API, using \texttt{gpt-3.5-turbo-0301}. Additionally, for $\text{Llama2}_\text{13b}$, we utilize the model parameters provided in the original code. We set the generation temperature to 0.3. Unless otherwise specified, we set the number of generations $k$ as 3. We use the \texttt{gpt-3.5-turbo-0301} to serve as our relation discriminator scoring agents to value. To ensure the reliability of the results, we conduct five rounds of experiments for each dataset, taking their average scores as the evaluation results. The prompts for CoT and PS are from \citet{wei2022chain} and \citet{Wang2023PlanandSolvePI} as a comparison to our proposed framework. 
For evaluation metrics, we utilize exact match (EM) and Accuracy (Acc) in our experiments. More details refers to Appendix~\ref{sec:Evaluation}.

\section{Experiments}

\subsection{Main Results}

{
\setlength{\parindent}{0cm}
\textbf{ERA-CoT outperforms all baselines in all benchmarks on $\text{Llama2}_\text{13B}$ and 5 out of 6 benchmarks on GPT3.5.} 
From Table~\ref{tab:main_result}, we can observe that our method shows great capability in three categories of reasoning problems.
For instance, ERA-CoT achieves an average improvement of 3.8\% on GPT3.5. 
This indicates that through entity relation knowledge, the LLMs could make better predictions and enhance their performance.
}


\vspace{5pt}

{
\setlength{\parindent}{0cm}

\textbf{Commonsense reasoning.} For the StrategyQA and CommonsenseQA datasets, ERA-CoT shows an average improvement of approximately \textbf{+6.1\%} compared to CoT. Although CoT-SC@5 could improve the performance of CoT, it still has a disparity of average scores with ERA-CoT on GPT3.5 (71.7\% vs. 77.3\%). It is worth noting that StrategyQA gets a performance drop compared to Vanilla LM, which may be caused by irrelevant text or incorrect reasoning chains. ERA-CoT mitigates the impact of irrelevant text by controlling the reasoning process based on entity relations. It performs a similar outcome on the $\text{Llama}_\text{13B}$ when we apply entity relation as a prompt to instruct the model. On the StrategyQA dataset, the result achieved an improvement of \textbf{+3.1\%} compared to RE2. Similar improvements are observed on the CommonsenseQA dataset, which indicates the ERA-CoT method could potentially result in very strong performances.

}

\begin{table}[tp]
\large
	 \resizebox{\linewidth}{!}{
\begin{tabular}{llcccc}
\toprule
\multirow{2}{*}{\textbf{Model}} & \multirow{2}{*}{\textbf{Datasets}} & \multirow{2}{*}{\textbf{Only EE}} & \multirow{2}{*}{\textbf{EE+ERE}} & \multirow{2}{*}{\textbf{EE+ERI}} & \multirow{2}{*}{\textbf{ERA-CoT}} \\
                                &                                    &                               &                                  &                                  &                                     \\ \midrule
\hspace{0.3cm}\multirow{6}{*}{\rotatebox{90}{GPT3.5}}        & \multicolumn{1}{l}{StrategyQA}    & 65.2                          & 67.9                             & 67.3                                                                  & 69.4                              \\
                                & CSQA          & 77.9                          & 80.5                             & 81.1                                                                  & 83.2                              \\
                                & LogiQA        & 37.2                          & 41.5                             & 42.1                                                              & 45.2                              \\
                                & HotpotQA      & 53.5                          & 55.8                             & 55.9                                                              & 58.4                              \\
                                & 2WikiMHQA     & 64.2                          & 68.1                             & 67.2                                                              & 70.2                              \\
                                & GSM8K         & 77.5                          & 78.6                             & 77.8                                                               & 78.2                              \\ \midrule
\hspace{0.3cm}\multirow{6}{*}{\rotatebox{90}{$\text{Llama}_\text{13B}$}}      & \multicolumn{1}{l}{StrategyQA}    & 57.1                          & 57.7                             & 58.2                                                            & 60.5                              \\
                                & \multicolumn{1}{l}{CSQA}          & 65.7                          & 68.9                             & 68.1                                                              & 72.6                              \\
                                & \multicolumn{1}{l}{LogiQA}        & 31.9                          & 32.4                             & 33.7                                                              & 35.5                              \\
                                & \multicolumn{1}{l}{HotpotQA}      & 35.8                          & 36.6                             & 36.9                                                                & 39.2                              \\
                                & \multicolumn{1}{l}{2WikiMHQA}     & 34.4                          & 34.9                             & 35.2                                   & 38.9                              \\
                                & \multicolumn{1}{l}{GSM8K}         & 22.9                          & 24.1                             & 24.7                              & 24.5                              \\ \bottomrule 
\end{tabular}}
    \caption{Performance comparisons upon different combinations and settings of entities relation steps. 
    }
\label{tab:side_result}
\end{table}

\vspace{5pt}

{
\setlength{\parindent}{0cm}

\textbf{Logical Reasoning.} Logical Reasoning contains more implicit relations. According to Table~\ref{tab:main_result}, ERA-CoT evaluates three logical datasets: LogiQA, HotpotQA, and 2WikiMHQA. The results on these three datasets demonstrate an average \textbf{+5.1\%} improvement on GPT3.5, highlighting the effectiveness of this method compared to other baselines.
Meanwhile, \textbf{ERA-CoT exhibits a better performance increment in logical reasoning tasks} that surpasses commonsense reasoning tasks and mathematical reasoning tasks, suggesting that our approach may be more suitable for tasks involving relational reasoning. On a small-scale LLM, ERA-CoT $\text{Llama}_{13B}$ achieves similar performance, indicating that ERA-CoT may have better capabilities in long-text comprehension and entity logical reasoning. Compared to the recent competitive work RE2, our method shows an average relative improvement of \textbf{5.2\%} in logical reasoning.

}

\vspace{5pt}

{
\setlength{\parindent}{0cm}

\textbf{Mathematical Reasoning.} The mathematical reasoning ability of ERA-CoT is evaluated on the GSM8K dataset. Compared to other methods, ERA-CoT outperforms most baselines on GSM8K, falling slightly behind Complex-CoT. Our approach primarily relies on the analysis of contextual entity relationships to assist the model in understanding the problems. As GSM8K involves natural language-formulated questions, this process potentially addresses some errors in the analysis of relational chains in CoT.

}

\begin{table*}[]
\renewcommand{\arraystretch}{1.2}
	 \resizebox{\linewidth}{!}{		
\scriptsize          
\begin{tabular}{lllllllllll}
\toprule
\multicolumn{1}{c}{\multirow{2}{*}{\textbf{Model}}} & \multicolumn{1}{c}{\multirow{2}{*}{\textbf{Dateset}}} & \multicolumn{4}{c}{\textbf{validation}} & \multicolumn{4}{c}{\textbf{w/o validation}} \\ \cmidrule{3-10} 
\multicolumn{1}{c}{} & \multicolumn{1}{c}{} & $k$@1 & $k$@3 & $k$@5 & $k$@10 & $k$@1 & \multicolumn{1}{c}{$k$@3} & \multicolumn{1}{c}{$k$@5} & \multicolumn{1}{c}{$k$@10} \\ \midrule
 & StrategyQA & 66.4 & 71.4 & 72.8 & 73.1 & 65.2 & 65.9 & 70.8 & 71.9 \\
 & CSQA & 79.5 & 83.2 & 84.8 & 85.6 & 78.5 & 80.2 & 83.4 & 84.8 \\
\multicolumn{1}{c}{GPT3.5} & LogiQA & 41.0 & 45.2 & 46.5 & 47.5 & 39.8 & 43.3 & 44.4 & 45.1 \\
 & HotpotQA & 54.4 & 58.4 & 61.2 & 63.5 & 53.2 & 53.9 & 57.1 & 58.5 \\
 & 2WikiMHQA & 67.3 & 70.2 & 71.1 & 72.2 & 66.5 & 68.7 & 67.3 & 70.9 \\
 & GSM8K & 75.8 & 79.5 & 80.2 & 80.1 & 74.9 & 76.0 & 77.1 & 77.3 \\ \hline
 & StrategyQA & 58.2 & 61.5 & 65.9 & 64.9 & 57.5 & 57.9 & 61.3 & 60.5 \\
 & CSQA & 68.4 & 72.6 & 71.3 & 70.1 & 68.1 & 69.0 & 70.3 & 71.6 \\
\multicolumn{1}{c}{$\text{Llama2}_\text{13B}$} & LogiQA & 33.6 & 35.5 & 37.8 & 40.5 & 31.7 & 34.1 & 35.7 & 38.4 \\
 & HotpotQA & 37.6 & 39.2 & 40.9 & 42.0 & 36.9 & 38.1 & 41.2 & 41.0 \\
 & 2WikiMHQA & 36.0 & 38.9 & 39.5 & 41.1 & 35.5 & 37.2 & 38.1 & 38.9 \\
 & GSM8K & 22.1 & 24.5 & 26.9 & 27.3 & 21.8 & 24.0 & 25.6 & 26.9 \\ \bottomrule
\end{tabular}}
 \caption{The impact of different numbers and relationship discrimination step of implicit relations on model accuracy, $k$ indicates the number of implicit relations reasoning by model.}
\label{tab:Implicit Relationships}
\end{table*}

\begin{figure}[tp]
    \centering
    \includegraphics[width=0.98\linewidth]{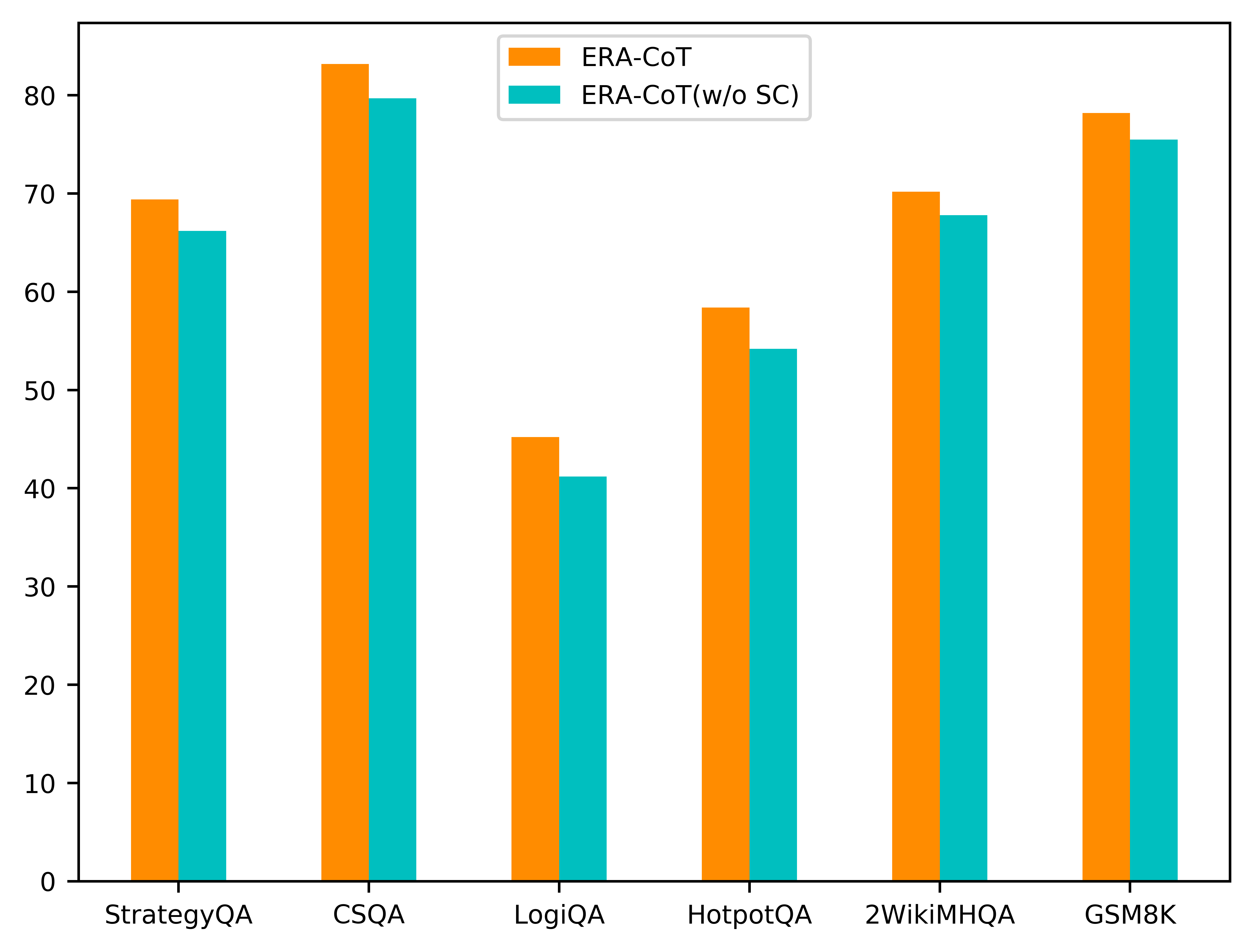}
    \caption{Comparison on the use of SC for the first two steps. Evaluation on the final performance.}
    \vspace{-0.15cm}
    \label{fig:sc}
\end{figure}

\subsection{Ablation Studies}

ERA-CoT includes multiple processes for handling entities and relationships. We combine various steps to assess the impact of entity extraction and relationship inference on model performance. We categorize the process into the following situations:

\begin{itemize}[leftmargin=1em]
    \item \textit{Only EE}: This variant represents only entity extraction, involving the use of large models for named entity recognition.
    \item \textit{EE+ERE}: This setting indicates simultaneous extraction of entities and explicit relationship extraction. After completing the first two steps of ERA-CoT, this process directly proceeds to question prediction using the answers from the output of the initial two steps.
    \item \textit{EE+ERI}: It implies that after entity extraction, we directly infer relationships between entities and make answer predictions based on the results of entity extraction and relationship inference.
\end{itemize}

\vspace{5pt}

{
\setlength{\parindent}{0cm}

\textbf{Entity extraction, relationship extraction and inference are effective for answering questions. }
Table~\ref{tab:side_result} demonstrates the positive impact of entity relationships on task inference. Performance of \textit{Only EE} declines on multiple datasets compared to ERA-CoT when there is no relationship extraction and inference. After entity extraction based on our prompts, task predictions show significant improvement.   On GPT3.5, the average performance for each task increased by \textbf{+2.0\%}, and on $\text{Llama}_\text{13B}$ it increased by \textbf{+1.1\%}. Additionally, direct relationship inference enhances the model's answer prediction performance. However, as it does not undergo relationship extraction, the inferred relationships lack based relations of entities, resulting in less competitive evaluation performance compared to ERA-CoT. Additionally, for mathematical reasoning, the performance of \textit{EE+ERE} and \textit{EE+ERI} are close to ERA-CoT. This may be attributed to the simplicity of relationships in mathematical tasks, making them easy to extract or identify.

}

\vspace{5pt}

{
\setlength{\parindent}{0cm}

\textbf{Effectiveness of Self-Consistency.} From Figure~\ref{fig:sc}, we can observe that 
removing SC resulted in an average performance decrease by an average of \textbf{-3.2\%} for the ERA-CoT method on GPT3.5. This result highlights the necessity of integrating self-consistency in the first two steps of this process.

}


\subsection{Analysis on Implicit Relationship}

In the process of deducing implicit relationships, we need to identify $k$ relationships between each pair of entities. Subsequently, these $k$ relationships are scored based on the contextual meaning, where higher scores indicate more reliable relationships, and lower scores suggest a lower likelihood of the relationship. By setting a reasonable threshold, implicit relationships with scores surpassing this threshold are retained for the final question prediction. We investigated the impact of the value of $k$ on model performance. Specifically, we conducted experiments on the ERA-CoT dataset using the GPT and $\text{Llama2}_\text{13B}$ models, evaluating their performance on different datasets.

\vspace{5pt}

{
\setlength{\parindent}{0cm}

\textbf{A reasonable number of implicit relationships contributes to a better understanding of the context.} The experimental results are shown in Table~\ref{tab:Implicit Relationships}. As the number of implicit relationships increases, the model's accuracy tends to improve. This indicates that discovering more implicit relationships enhances the model's reasoning ability, ultimately increasing the likelihood of correct answer predictions. Specifically, when $k$ is relatively small, the model's accuracy significantly improves. However, an excessively large number may lead to hallucinations. As $k$ continues to increase, the improvement in accuracy becomes smaller, and there may even be a slight decline in accuracy. This is because inferring too many implicit relationships may lead to illusion effects, affecting the model's final reasoning judgment. Therefore, considering the balance between model effectiveness and complexity, we found that setting $k$ to 3 results in better accuracy for the model. Additionally, in complex relationship scenarios, this choice does not significantly increase the complexity of the approach.
}

\vspace{5pt}

{
\setlength{\parindent}{0cm}

\textbf{Relation discrimination steps help eliminate incorrect relationship pairs.} We investigated whether performing relation discrimination steps during model accuracy evaluation plays an important role. In the discrimination step, the model scores all implicit relationships and filters out those with scores below $v_{th}$. The results show that models with discrimination steps exhibit higher accuracy compared to models without discrimination steps. This process helps enhance the correctness and robustness of the model's reasoning, explaining the overall higher accuracy of models with discrimination steps.
}

\vspace{5pt}

{
\setlength{\parindent}{0cm}

\textbf{Small LLMs still possess the ability to identify implicit relationships.} Our experiments on $\text{Llama2}_\text{13B}$ indicate that as $k$ increases from 1 to 5, the model shows an average improvement of \textbf{+4.4\%}. However, small models have a weaker understanding of the text, and as the number of inferred implicit relationships continues to increase, small models are more prone to generating incorrect relationships, leading to unstable performance. 

}

\subsection{Low relations density sentences analysis}

In simple sentences and basic questions, the limited relationships make it challenging to learn implicit relationships from the context. Nevertheless, ERA-CoT can still provide some degree of assistance. Compared to some knowledge-intensive tasks, the performance improvement might be less noticeable.

A typical example is the CommonsenseQA dataset, which mainly focuses on short, single-sentence questions involving commonsense reasoning, characterized by fewer entities and relationships. On CSQA, we achieve state-of-the-art results, showing a clear advantage over other baseline methods. Furthermore, as shown in Table~\ref{tab:Implicit Relationships}, while the effectiveness of our approach doesn't expand as significantly with increased implicit relationship reasoning compared to datasets like HotpotQA or LogiQA, there is still a noticeable improvement from 1 to 3. This indicates that even simple sentences can contain some in-depth entity relationships worth exploring.

\begin{table}[]
	 \resizebox{\linewidth}{!}{
\begin{tabular}{@{}cc|llll@{}}
\toprule
 \textbf{Task} & \multicolumn{1}{c|}{\textbf{Dateset}} & \textbf{\textit{En.}} & \textbf{\textit{Ex.}} & \textbf{\textit{Im.}} & \textbf{\textit{An.}} \\ \midrule
\multirow{2}{*}{\textbf{\begin{tabular}[c]{@{}c@{}}Commonsense \\ Reasoning\end{tabular}}} & StrategyQA & 7\% & 21\% & 23\% & 15\% \\
 & CSQA & 6\% & 16\% & 16\% & 10\% \\ \midrule
\multirow{3}{*}{\textbf{\begin{tabular}[c]{@{}l@{}}Logical\\ Reasoning\end{tabular}}} & LogiQA & 10\% & 21\% & 32\% & 3\% \\
 & HotpotQA & 7\% & 16\% & 38\% & 10\% \\
 & 2WikiMHQA & 8\% & 22\% & 35\% & 13\% \\ \midrule
\textbf{\begin{tabular}[c]{@{}c@{}}Mathematical\\ Reasoning\end{tabular}} & GSM8K & 2\% & 10\% & 21\% & 12\% \\ \bottomrule
\end{tabular}}
\caption{Error categories per dataset.
Multiple categories are allowed for each example.}
\label{tab:Error analysis}
\end{table}

\subsection{Error analysis}

We manually analyzed 100 errors in each dataset and categorized them into following four categories: (i) \textbf{Entity Extraction Errors \textit{(En.)}} – failure to recognize all entities in the text; (ii) \textbf{Explicit Relationship Extraction Errors \textit{(Ex.)}} – entities extracted correctly, but failing to extract all explicit relationships in the text or extracting non-existing explicit relationships; (iii) \textbf{Implicit Relationship Inference Errors \textit{(Im.)}} – correct extraction of entities and explicit relationships, but inferring non-existing implicit relationships; and (iv) \textbf{Answer Errors \textit{(An.)}} – correct inference of relationships by \textit{ERI}, but providing an incorrect answer.

\vspace{5pt}

{
\setlength{\parindent}{0cm}

\textbf{Implicit relationship inference is prone to errors.} Table~\ref{tab:Error analysis} shows the error category results for each dataset. Considering the error categories, the probability of \textit{Im.} is consistently the highest, while \textit{En.} has a relatively lower probability. This suggests that the inference of implicit relationships has the most significant impact on the model's accuracy.
}

\vspace{5pt}

{
\setlength{\parindent}{0cm}
\textbf{Error rates are also influenced by dataset characteristics. }We observe that for common-sense reasoning datasets, error rates for \textit{Ex.} and \textit{Im.} are close. This is attributed to the model potentially misjudging relationships between entities due to an incomplete understanding of common-sense knowledge.
}

For logical reasoning datasets, \textit{Ex.} is higher compared to other datasets. This is because these datasets typically involve longer texts with more relationships, requiring the model to have a higher level of relationship extraction. However, compared to other errors, the error rate of \textit{An.} is relatively small. This implies that if the relationships between entities are correctly inferred, the model is likely to answer questions correctly. This indicates that accurate relationship inference is highly beneficial for correctly answering questions in this type of dataset.

For mathematical reasoning datasets, implicit inference contributes to improving dataset accuracy, but \textit{An.} should not be ignored. This is because even if relationship inference is correct, errors in calculations may still lead to incorrect answers.

\section{Conclusions}

In this paper, we introduce the method of ERA-CoT to address open-domain question answering and knowledge reasoning tasks. 
By leveraging inference and detection of entity relationships, this method exhibits remarkable performance in tasks with lengthy texts or those involving numerous and complex entity relationships. 
Extensive experiments demonstrate the superiority of the proposed method in various reasoning categories. We hope that our method can be applied to enhance the performance of LLMs in diverse domains.

\section*{Limitation}

Our work has limitations in certain scenarios. First, ERA-CoT relies on context analysis through the extraction of relationships and entities, so its performance improvement is not significant in tasks with fewer entity relationships, such as symbolic reasoning tasks. Additionally, due to the various relationships between entities, even after relationship extraction and inference, there may still be some relationships that the model fails to correctly infer or extract. This could result in missing relationships during result predictions, affecting the accuracy of predictions. In future work, we aim to address these two issues. We plan to explore whether the model can effectively understand the internal structure of entities and conduct a more in-depth analysis of entity relationships to enhance the effectiveness of LLM prompting.

\section*{Ethics Statement}

The datasets we used are sourced from the current public datasets. The prompts we employed do not collect or utilize personal information or information from other individuals. Furthermore, they do not contain any sensitive words or oppose any individual or group. Our work only extracts the entity and relation from these datasets to predict the answer to tasks.  Our method strictly adheres to the license and policies of
 released LLMs and publicly available datasets, and our work could be further integrated with other methods.

\bibliography{anthology,custom}

\begin{thebibliography}{49}
\expandafter\ifx\csname natexlab\endcsname\relax\def\natexlab#1{#1}\fi

\bibitem[{Ashok and Lipton(2023)}]{ashok2023promptner}
Dhananjay Ashok and Zachary~C. Lipton. 2023.
\newblock \href {http://arxiv.org/abs/2305.15444} {Promptner: Prompting for named entity recognition}.

\bibitem[{Brown et~al.(2020)Brown, Mann, Ryder, Subbiah, Kaplan, Dhariwal, Neelakantan, Shyam, Sastry, Askell et~al.}]{brown2020language}
Tom Brown, Benjamin Mann, Nick Ryder, Melanie Subbiah, Jared~D Kaplan, Prafulla Dhariwal, Arvind Neelakantan, Pranav Shyam, Girish Sastry, Amanda Askell, et~al. 2020.
\newblock Language models are few-shot learners.
\newblock \emph{Advances in neural information processing systems}, 33:1877--1901.

\bibitem[{Cheng and Erk(2020)}]{cheng2020attending}
Pengxiang Cheng and Katrin Erk. 2020.
\newblock Attending to entities for better text understanding.
\newblock In \emph{Proceedings of the AAAI conference on artificial intelligence}, volume~34, pages 7554--7561.

\bibitem[{Chowdhery et~al.(2023)Chowdhery, Narang, Devlin, Bosma, Mishra, Roberts, Barham, Chung, Sutton, Gehrmann et~al.}]{chowdhery2023palm}
Aakanksha Chowdhery, Sharan Narang, Jacob Devlin, Maarten Bosma, Gaurav Mishra, Adam Roberts, Paul Barham, Hyung~Won Chung, Charles Sutton, Sebastian Gehrmann, et~al. 2023.
\newblock Palm: Scaling language modeling with pathways.
\newblock \emph{Journal of Machine Learning Research}, 24(240):1--113.

\bibitem[{Chu et~al.(2023)Chu, Chen, Chen, Yu, He, Wang, Peng, Liu, Qin, and Liu}]{chu2023survey}
Zheng Chu, Jingchang Chen, Qianglong Chen, Weijiang Yu, Tao He, Haotian Wang, Weihua Peng, Ming Liu, Bing Qin, and Ting Liu. 2023.
\newblock A survey of chain of thought reasoning: Advances, frontiers and future.
\newblock \emph{arXiv preprint arXiv:2309.15402}.

\bibitem[{Cobbe et~al.(2021)Cobbe, Kosaraju, Bavarian, Chen, Jun, Kaiser, Plappert, Tworek, Hilton, Nakano et~al.}]{cobbe2021training}
Karl Cobbe, Vineet Kosaraju, Mohammad Bavarian, Mark Chen, Heewoo Jun, Lukasz Kaiser, Matthias Plappert, Jerry Tworek, Jacob Hilton, Reiichiro Nakano, et~al. 2021.
\newblock Training verifiers to solve math word problems.
\newblock \emph{arXiv preprint arXiv:2110.14168}.

\bibitem[{Das et~al.(2022)Das, Katiyar, Passonneau, and Zhang}]{das2022container}
Sarkar Snigdha~Sarathi Das, Arzoo Katiyar, Rebecca~J Passonneau, and Rui Zhang. 2022.
\newblock Container: Few-shot named entity recognition via contrastive learning.
\newblock In \emph{Proceedings of the 60th Annual Meeting of the Association for Computational Linguistics (Volume 1: Long Papers)}, pages 6338--6353.

\bibitem[{Fu et~al.(2023)Fu, Peng, Sabharwal, Clark, and Khot}]{fu2023complexitybased}
Yao Fu, Hao Peng, Ashish Sabharwal, Peter Clark, and Tushar Khot. 2023.
\newblock \href {https://openreview.net/forum?id=yf1icZHC-l9} {Complexity-based prompting for multi-step reasoning}.
\newblock In \emph{The Eleventh International Conference on Learning Representations}.

\bibitem[{Ganguli et~al.(2023)Ganguli, Askell, Schiefer, Liao, Luko{\v{s}}i{\=u}t{\.e}, Chen et~al.}]{ganguli2302capacity}
D~Ganguli, A~Askell, N~Schiefer, T~Liao, K~Luko{\v{s}}i{\=u}t{\.e}, A~Chen, et~al. 2023.
\newblock The capacity for moral self-correction in large language models.
\newblock \emph{arXiv preprint arXiv:2302.07459}.

\bibitem[{Geva et~al.(2021)Geva, Khashabi, Segal, Khot, Roth, and Berant}]{geva2021did}
Mor Geva, Daniel Khashabi, Elad Segal, Tushar Khot, Dan Roth, and Jonathan Berant. 2021.
\newblock Did aristotle use a laptop? a question answering benchmark with implicit reasoning strategies.
\newblock \emph{Transactions of the Association for Computational Linguistics}, 9:346--361.

\bibitem[{Han et~al.(2019)Han, Gao, Yao, Ye, Liu, and Sun}]{han-etal-2019-opennre}
Xu~Han, Tianyu Gao, Yuan Yao, Deming Ye, Zhiyuan Liu, and Maosong Sun. 2019.
\newblock \href {https://doi.org/10.18653/v1/D19-3029} {{O}pen{NRE}: An open and extensible toolkit for neural relation extraction}.
\newblock In \emph{Proceedings of the 2019 Conference on Empirical Methods in Natural Language Processing and the 9th International Joint Conference on Natural Language Processing (EMNLP-IJCNLP): System Demonstrations}, pages 169--174, Hong Kong, China. Association for Computational Linguistics.

\bibitem[{Ho et~al.(2020)Ho, Nguyen, Sugawara, and Aizawa}]{ho2020constructing}
Xanh Ho, Anh-Khoa~Duong Nguyen, Saku Sugawara, and Akiko Aizawa. 2020.
\newblock Constructing a multi-hop qa dataset for comprehensive evaluation of reasoning steps.
\newblock In \emph{Proceedings of the 28th International Conference on Computational Linguistics}, pages 6609--6625.

\bibitem[{Hoffmann et~al.(2022)Hoffmann, Borgeaud, Mensch, Buchatskaya, Cai, Rutherford, Casas, Hendricks, Welbl, Clark et~al.}]{hoffmann2022training}
Jordan Hoffmann, Sebastian Borgeaud, Arthur Mensch, Elena Buchatskaya, Trevor Cai, Eliza Rutherford, Diego de~Las Casas, Lisa~Anne Hendricks, Johannes Welbl, Aidan Clark, et~al. 2022.
\newblock Training compute-optimal large language models.
\newblock \emph{arXiv preprint arXiv:2203.15556}.

\bibitem[{Jimenez~Gutierrez et~al.(2022)Jimenez~Gutierrez, McNeal, Washington, Chen, Li, Sun, and Su}]{jimenez-gutierrez-etal-2022-thinking}
Bernal Jimenez~Gutierrez, Nikolas McNeal, Clayton Washington, You Chen, Lang Li, Huan Sun, and Yu~Su. 2022.
\newblock \href {https://doi.org/10.18653/v1/2022.findings-emnlp.329} {Thinking about {GPT}-3 in-context learning for biomedical {IE}? think again}.
\newblock In \emph{Findings of the Association for Computational Linguistics: EMNLP 2022}, pages 4497--4512, Abu Dhabi, United Arab Emirates. Association for Computational Linguistics.

\bibitem[{Kojima et~al.(2022)Kojima, Gu, Reid, Matsuo, and Iwasawa}]{kojima2022large}
Takeshi Kojima, Shixiang~Shane Gu, Machel Reid, Yutaka Matsuo, and Yusuke Iwasawa. 2022.
\newblock Large language models are zero-shot reasoners.
\newblock \emph{Advances in neural information processing systems}, 35:22199--22213.

\bibitem[{Lazaridou et~al.(2022)Lazaridou, Gribovskaya, Stokowiec, and Grigorev}]{lazaridou2022internet}
Angeliki Lazaridou, Elena Gribovskaya, Wojciech Stokowiec, and Nikolai Grigorev. 2022.
\newblock Internet-augmented language models through few-shot prompting for open-domain question answering.
\newblock \emph{arXiv preprint arXiv:2203.05115}.

\bibitem[{Li et~al.(2023)Li, Wang, and Ke}]{li2023revisiting}
Guozheng Li, Peng Wang, and Wenjun Ke. 2023.
\newblock Revisiting large language models as zero-shot relation extractors.
\newblock In \emph{Findings of the Association for Computational Linguistics: EMNLP 2023}, pages 6877--6892.

\bibitem[{Li et~al.(2022)Li, Zhang, and Zhao}]{li2022self}
Junlong Li, Zhuosheng Zhang, and Hai Zhao. 2022.
\newblock Self-prompting large language models for open-domain qa.
\newblock \emph{arXiv preprint arXiv:2212.08635}.

\bibitem[{Li and Du(2023)}]{li2023leveraging}
Ruosen Li and Xinya Du. 2023.
\newblock Leveraging structured information for explainable multi-hop question answering and reasoning.
\newblock In \emph{Findings of the Association for Computational Linguistics: EMNLP 2023}, pages 6779--6789.

\bibitem[{Liu et~al.(2021)Liu, Cui, Liu, Huang, Wang, and Zhang}]{liu2021logiqa}
Jian Liu, Leyang Cui, Hanmeng Liu, Dandan Huang, Yile Wang, and Yue Zhang. 2021.
\newblock Logiqa: a challenge dataset for machine reading comprehension with logical reasoning.
\newblock In \emph{Proceedings of the Twenty-Ninth International Conference on International Joint Conferences on Artificial Intelligence}, pages 3622--3628.

\bibitem[{Luo et~al.(2020{\natexlab{a}})Luo, Su, and Yu}]{luo2020bert}
Da~Luo, Jindian Su, and Shanshan Yu. 2020{\natexlab{a}}.
\newblock A bert-based approach with relation-aware attention for knowledge base question answering.
\newblock In \emph{2020 International Joint Conference on Neural Networks (IJCNN)}, pages 1--8. IEEE.

\bibitem[{Luo et~al.(2020{\natexlab{b}})Luo, Xiao, and Zhao}]{luo2020hierarchical}
Ying Luo, Fengshun Xiao, and Hai Zhao. 2020{\natexlab{b}}.
\newblock Hierarchical contextualized representation for named entity recognition.
\newblock In \emph{Proceedings of the AAAI conference on artificial intelligence}, volume~34, pages 8441--8448.

\bibitem[{Lyu and Chen(2021)}]{lyu-chen-2021-relation}
Shengfei Lyu and Huanhuan Chen. 2021.
\newblock \href {https://doi.org/10.18653/v1/2021.findings-acl.34} {Relation classification with entity type restriction}.
\newblock In \emph{Findings of the Association for Computational Linguistics: ACL-IJCNLP 2021}, pages 390--395, Online. Association for Computational Linguistics.

\bibitem[{Malmasi et~al.(2022)Malmasi, Fang, Fetahu, Kar, and Rokhlenko}]{malmasi2022multiconer}
Shervin Malmasi, Anjie Fang, Besnik Fetahu, Sudipta Kar, and Oleg Rokhlenko. 2022.
\newblock Multiconer: A large-scale multilingual dataset for complex named entity recognition.
\newblock In \emph{Proceedings of the 29th International Conference on Computational Linguistics}, pages 3798--3809.

\bibitem[{Min et~al.(2022)Min, Lewis, Zettlemoyer, and Hajishirzi}]{min-etal-2022-metaicl}
Sewon Min, Mike Lewis, Luke Zettlemoyer, and Hannaneh Hajishirzi. 2022.
\newblock \href {https://doi.org/10.18653/v1/2022.naacl-main.201} {{M}eta{ICL}: Learning to learn in context}.
\newblock In \emph{Proceedings of the 2022 Conference of the North American Chapter of the Association for Computational Linguistics: Human Language Technologies}, pages 2791--2809, Seattle, United States. Association for Computational Linguistics.

\bibitem[{Moscato et~al.(2023)Moscato, Postiglione, and Sperl{\'\i}}]{moscato2023few}
Vincenzo Moscato, Marco Postiglione, and Giancarlo Sperl{\'\i}. 2023.
\newblock Few-shot named entity recognition: definition, taxonomy and research directions.
\newblock \emph{ACM Transactions on Intelligent Systems and Technology}, 14(5):1--46.

\bibitem[{Moslem et~al.(2023)Moslem, Haque, and Way}]{moslem2023adaptive}
Yasmin Moslem, Rejwanul Haque, and Andy Way. 2023.
\newblock Adaptive machine translation with large language models.
\newblock \emph{arXiv preprint arXiv:2301.13294}.

\bibitem[{Narang et~al.(2023)Narang, Chowdhery, and Zhou}]{narangself}
Sharan Narang, Aakanksha Chowdhery, and Denny Zhou. 2023.
\newblock Self-consistency improves chain of thought reasoning in language models.
\newblock In \emph{Proceedings of the Eleventh International Conference on Learning Representations}.

\bibitem[{OpenAI(2023)}]{openai2023gpt4}
OpenAI. 2023.
\newblock \href {http://arxiv.org/abs/2303.08774} {Gpt-4 technical report}.

\bibitem[{Robinson et~al.(2022)Robinson, Rytting, and Wingate}]{robinson2022leveraging}
Joshua Robinson, Christopher~Michael Rytting, and David Wingate. 2022.
\newblock Leveraging large language models for multiple choice question answering.
\newblock \emph{arXiv preprint arXiv:2210.12353}.

\bibitem[{Shi et~al.(2023)Shi, Chen, Misra, Scales, Dohan, Chi, Sch{\"a}rli, and Zhou}]{shi2023large}
Freda Shi, Xinyun Chen, Kanishka Misra, Nathan Scales, David Dohan, Ed~H Chi, Nathanael Sch{\"a}rli, and Denny Zhou. 2023.
\newblock Large language models can be easily distracted by irrelevant context.
\newblock In \emph{International Conference on Machine Learning}, pages 31210--31227. PMLR.

\bibitem[{Tabassum et~al.(2020)Tabassum, Maddela, Xu, and Ritter}]{tabassum2020code}
Jeniya Tabassum, Mounica Maddela, Wei Xu, and Alan Ritter. 2020.
\newblock Code and named entity recognition in stackoverflow.
\newblock In \emph{Proceedings of the 58th Annual Meeting of the Association for Computational Linguistics}, pages 4913--4926.

\bibitem[{Taill{\'e} et~al.(2020)Taill{\'e}, Guigue, and Gallinari}]{taille2020contextualized}
Bruno Taill{\'e}, Vincent Guigue, and Patrick Gallinari. 2020.
\newblock Contextualized embeddings in named-entity recognition: An empirical study on generalization.
\newblock In \emph{Advances in Information Retrieval: 42nd European Conference on IR Research, ECIR 2020, Lisbon, Portugal, April 14--17, 2020, Proceedings, Part II 42}, pages 383--391. Springer.

\bibitem[{Talmor et~al.(2019)Talmor, Herzig, Lourie, and Berant}]{talmor2019commonsenseqa}
Alon Talmor, Jonathan Herzig, Nicholas Lourie, and Jonathan Berant. 2019.
\newblock Commonsenseqa: A question answering challenge targeting commonsense knowledge.
\newblock In \emph{Proceedings of the 2019 Conference of the North American Chapter of the Association for Computational Linguistics: Human Language Technologies, Volume 1 (Long and Short Papers)}, pages 4149--4158.

\bibitem[{Touvron et~al.(2023)Touvron, Lavril, Izacard, Martinet, Lachaux, Lacroix, Rozi{\`e}re, Goyal, Hambro, Azhar et~al.}]{touvron2023llama}
Hugo Touvron, Thibaut Lavril, Gautier Izacard, Xavier Martinet, Marie-Anne Lachaux, Timoth{\'e}e Lacroix, Baptiste Rozi{\`e}re, Naman Goyal, Eric Hambro, Faisal Azhar, et~al. 2023.
\newblock Llama: Open and efficient foundation language models.
\newblock \emph{arXiv preprint arXiv:2302.13971}.

\bibitem[{Vilar et~al.(2022)Vilar, Freitag, Cherry, Luo, Ratnakar, and Foster}]{vilar2022prompting}
David Vilar, Markus Freitag, Colin Cherry, Jiaming Luo, Viresh Ratnakar, and George Foster. 2022.
\newblock Prompting palm for translation: Assessing strategies and performance.
\newblock \emph{arXiv preprint arXiv:2211.09102}.

\bibitem[{Wan et~al.(2023)Wan, Cheng, Mao, Liu, Song, Li, and Kurohashi}]{wan2023gpt}
Zhen Wan, Fei Cheng, Zhuoyuan Mao, Qianying Liu, Haiyue Song, Jiwei Li, and Sadao Kurohashi. 2023.
\newblock Gpt-re: In-context learning for relation extraction using large language models.
\newblock \emph{arXiv preprint arXiv:2305.02105}.

\bibitem[{Wang et~al.(2023{\natexlab{a}})Wang, Xu, Lan, Hu, Lan, Lee, and Lim}]{Wang2023PlanandSolvePI}
Lei Wang, Wanyu Xu, Yihuai Lan, Zhiqiang Hu, Yunshi Lan, Roy Ka-Wei Lee, and Ee-Peng Lim. 2023{\natexlab{a}}.
\newblock \href {https://api.semanticscholar.org/CorpusID:258558102} {Plan-and-solve prompting: Improving zero-shot chain-of-thought reasoning by large language models}.
\newblock In \emph{Annual Meeting of the Association for Computational Linguistics}.

\bibitem[{Wang et~al.(2023{\natexlab{b}})Wang, Sun, Li, Ouyang, Wu, Zhang, Li, and Wang}]{wang2023gpt}
Shuhe Wang, Xiaofei Sun, Xiaoya Li, Rongbin Ouyang, Fei Wu, Tianwei Zhang, Jiwei Li, and Guoyin Wang. 2023{\natexlab{b}}.
\newblock Gpt-ner: Named entity recognition via large language models.
\newblock \emph{arXiv preprint arXiv:2304.10428}.

\bibitem[{Wang et~al.(2023{\natexlab{c}})Wang, Wei, Schuurmans, Le, Chi, Narang, Chowdhery, and Zhou}]{wang2023selfconsistency}
Xuezhi Wang, Jason Wei, Dale Schuurmans, Quoc~V Le, Ed~H. Chi, Sharan Narang, Aakanksha Chowdhery, and Denny Zhou. 2023{\natexlab{c}}.
\newblock \href {https://openreview.net/forum?id=1PL1NIMMrw} {Self-consistency improves chain of thought reasoning in language models}.
\newblock In \emph{The Eleventh International Conference on Learning Representations}.

\bibitem[{Wei et~al.(2022)Wei, Wang, Schuurmans, Bosma, Xia, Chi, Le, Zhou et~al.}]{wei2022chain}
Jason Wei, Xuezhi Wang, Dale Schuurmans, Maarten Bosma, Fei Xia, Ed~Chi, Quoc~V Le, Denny Zhou, et~al. 2022.
\newblock Chain-of-thought prompting elicits reasoning in large language models.
\newblock \emph{Advances in Neural Information Processing Systems}, 35:24824--24837.

\bibitem[{Wei et~al.(2023)Wei, Cui, Cheng, Wang, Zhang, Huang, Xie, Xu, Chen, Zhang, Jiang, and Han}]{wei2023zeroshot}
Xiang Wei, Xingyu Cui, Ning Cheng, Xiaobin Wang, Xin Zhang, Shen Huang, Pengjun Xie, Jinan Xu, Yufeng Chen, Meishan Zhang, Yong Jiang, and Wenjuan Han. 2023.
\newblock \href {http://arxiv.org/abs/2302.10205} {Zero-shot information extraction via chatting with chatgpt}.

\bibitem[{Xu et~al.(2023{\natexlab{a}})Xu, Tao, Shen, Xu, Xu, Long, and guang Lou}]{xu2023rereading}
Xiaohan Xu, Chongyang Tao, Tao Shen, Can Xu, Hongbo Xu, Guodong Long, and Jian guang Lou. 2023{\natexlab{a}}.
\newblock \href {http://arxiv.org/abs/2309.06275} {Re-reading improves reasoning in language models}.

\bibitem[{Xu et~al.(2023{\natexlab{b}})Xu, Zhu, Wang, and Zhang}]{xu-etal-2023-unleash}
Xin Xu, Yuqi Zhu, Xiaohan Wang, and Ningyu Zhang. 2023{\natexlab{b}}.
\newblock \href {https://doi.org/10.18653/v1/2023.sustainlp-1.13} {How to unleash the power of large language models for few-shot relation extraction?}
\newblock In \emph{Proceedings of The Fourth Workshop on Simple and Efficient Natural Language Processing (SustaiNLP)}, pages 190--200, Toronto, Canada (Hybrid). Association for Computational Linguistics.

\bibitem[{Yang et~al.(2018)Yang, Qi, Zhang, Bengio, Cohen, Salakhutdinov, and Manning}]{yang2018hotpotqa}
Zhilin Yang, Peng Qi, Saizheng Zhang, Yoshua Bengio, William Cohen, Ruslan Salakhutdinov, and Christopher~D Manning. 2018.
\newblock Hotpotqa: A dataset for diverse, explainable multi-hop question answering.
\newblock In \emph{Proceedings of the 2018 Conference on Empirical Methods in Natural Language Processing}, pages 2369--2380.

\bibitem[{Zhang et~al.(2023{\natexlab{a}})Zhang, Jimenez~Gutierrez, and Su}]{zhang-etal-2023-aligning}
Kai Zhang, Bernal Jimenez~Gutierrez, and Yu~Su. 2023{\natexlab{a}}.
\newblock \href {https://doi.org/10.18653/v1/2023.findings-acl.50} {Aligning instruction tasks unlocks large language models as zero-shot relation extractors}.
\newblock In \emph{Findings of the Association for Computational Linguistics: ACL 2023}, pages 794--812, Toronto, Canada. Association for Computational Linguistics.

\bibitem[{Zhang et~al.()Zhang, Han, Liu, Jiang, Sun, and Liu}]{zhangernie}
Zhengyan Zhang, Xu~Han, Zhiyuan Liu, Xin Jiang, Maosong Sun, and Qun Liu.
\newblock Ernie: Enhanced language representation with informative entities.

\bibitem[{Zhang et~al.(2023{\natexlab{b}})Zhang, Zhang, Li, and Smola}]{zhang2023automatic}
Zhuosheng Zhang, Aston Zhang, Mu~Li, and Alex Smola. 2023{\natexlab{b}}.
\newblock \href {https://openreview.net/forum?id=5NTt8GFjUHkr} {Automatic chain of thought prompting in large language models}.
\newblock In \emph{The Eleventh International Conference on Learning Representations}.

\bibitem[{Zhou and Chen(2022)}]{zhou-chen-2022-improved}
Wenxuan Zhou and Muhao Chen. 2022.
\newblock \href {https://aclanthology.org/2022.aacl-short.21} {An improved baseline for sentence-level relation extraction}.
\newblock In \emph{Proceedings of the 2nd Conference of the Asia-Pacific Chapter of the Association for Computational Linguistics and the 12th International Joint Conference on Natural Language Processing (Volume 2: Short Papers)}, pages 161--168, Online only. Association for Computational Linguistics.

\end{thebibliography}
\bibliographystyle{acl_natbib}

\appendix

\label{sec:appendix}

\section{Dataset Statistics}

 Table~\ref{tab:Dataset_analysis} provides detailed information about the data included in the experiment, where the sampled data are randomly selected from datasets, with a minimum of 1319 samples taken.

\begin{table}[H]
	 \resizebox{\linewidth}{!}{	
\begin{tabular}{llll}
\toprule
\textbf{Dataset}         & \textbf{Num.} & \textbf{Length} & \textbf{Domain}                 \\ \midrule
StrategyQA      & 1390                  & 12.3          & Commonsense Reasoning  \\
CSQA   & 3675                  & 18.5          & Commonsense Reasoning  \\ \midrule
LogiQA          & 1735                  & 103.8         & Logical Reasoning      \\
HotpotQA        & 3000                  & 114.2         & Logical Reasoning      \\
2WikiMHQA & 1539                   & 95.3          & Logical Reasoning      \\ \midrule
GMS8K           & 1319                  & 46.9          & Mathematical Reasoning \\ \bottomrule
\end{tabular}}

 \caption{
Dataset statistics, where ``Num.'' represents the number of sampled datasets, and ``Length'' is the number of average tokens in the sampled dataset.}
\label{tab:Dataset_analysis}
\end{table}

\section{Evaluation Metrics}
\label{sec:Evaluation}

We use accuracy and exact match as the evaluation metric for different datasets. Specifically, for datasets like StrategyQA, CSQA, and LogiQA that contain options, we utilize the accuracy based on whether the options match the standard answers. For problems like GSM8K, where the output is a number, we use regular expressions for exact match judgment of the answers. For datasets like 2WikiMQA that do not contain question options, we compare the output with answer alternatives and also use the exact match method for accuracy estimation. The same processing approach is adopted for different methods across these datasets.

\section{Experiment Cost}

We conduct experiments using the GPT3.5 API, utilizing the \texttt{gpt-3.5-turbo-0301} model, at a cost of \$0.002 per 1K tokens. In total, we spend \$529.

\section{Entity Types}

In the entity extraction step, we adopted the commonly used named entity types as indicated in the nltk official documentation. 

\begin{table}[h]
	 \resizebox{\linewidth}{!}{	
\begin{tabular}{ll}

\toprule
\multicolumn{1}{c}{\textbf{NE Type}} & \multicolumn{1}{c}{\textbf{Examples}}            \\ \midrule
ORGANIZATION                         & \textit{Georgia-Pacific Corp., WHO}              \\
PERSON                               & \textit{Eddy Bonte, President Obama}             \\
LOCATION                             & \textit{Murray River, Mount Everest}             \\
DATE                                 & \textit{June, 2008-06-29}                        \\
TIME                                 & \textit{two fifty a m, 1:30 p.m.}                \\
MONEY                                & \textit{175 million Canadian Dollars, GBP 10.40} \\
PERCENT                              & \textit{twenty pct, 18.75 \%}                    \\
FACILITY                             & \textit{Washington Monument, Stonehenge}         \\
GPE                                  & \textit{South East Asia, Midlothian}             \\ \bottomrule
\end{tabular}
}
\end{table}

If our entities, such as personal or place names, contain commas, we separate the entity with quotes "" to ensure correct information parsing across different steps.

\section{Prompting Template}

The following is the prompt statement used by ERA-CoT, guiding different steps that need to be processed with Chain-of-Thought. These prompts are proposed under a zero-shot setting, intended for resolution based on the initially provided context and query of the question.

\subsection{Prompt for Entities Extraction}
\begin{mybox3}{Entities Extraction}

Given a sentence, possible entities may include:[\textit{individuals, organizations, locations, ..., percentages}]. Find all entities based on the provided sentence.

\textbf{Sentence}: [Sentence $S$]

\textbf{Entities}: 

\end{mybox3}

\subsection{Prompt for Entities Relation Extraction}
\begin{mybox3}{Entities Relation Extraction}

Given a sentence, and all entities within the sentence. Extract all relationships between entities which directly stated in the sentence.

Every relationship stated as a triple: ($E_A$, $E_B$, $Relation$)

\textbf{Sentence}: [Sentence $S$]

\textbf{Entities}: [Entities List $\{E_i\}$]

\textbf{Relationships}:

\end{mybox3}

\subsection{Prompt for Entities Relation Inference}

\begin{mybox3}{Entities Relation Inference}

Given a sentence, all entities, and all explicit relationships within the sentence. Infer all possible implicit relationships between entities. For each pair of entities, infer up to [$k$] implicit relationships.

Every relationship stated as a triple: ($E_A$, $E_B$, $Relation$)

\textbf{Sentence}: [Sentence $S$]

\textbf{Entities}: [Entities List $\{E_i\}$]

\textbf{Explicit Relationships}:[Explicit Relationship List $\{R_e\}$]

\textbf{Implicit Relationships}:

\end{mybox3}

\subsection{Prompt for Relationship Discrimination}

\begin{mybox3}{Relationship Discrimination}

Given a sentence, and all uncertain relationships within the sentence. Score the confidence level of each relationship.

The confidence score ranges from 0 to 10, where a higher score indicates a higher likelihood of the relationship being correct.

Every relationship stated as a triple: ($E_A$, $E_B$, $Relation$)

\textbf{Sentence}: [Sentence $S$]

\textbf{Entities}: [Entities List $\{E_i\}$]

\textbf{Uncertain Relationships}:[Implicit Relationship List $\{R_i\}$]

\textbf{Scores}:

\end{mybox3}

\subsection{Prompt for Question Answering}

\begin{mybox3}{Question Answering}

Given a sentence, all entities and all relationships within the sentence. Answering the question.

Every relationship stated as a triple: ($E_A$, $E_B$, $Relation$)

\textbf{Sentence}: [Sentence $S$]

\textbf{Entities}: [Entities List $\{E_i\}$]

\textbf{Relationships}:[Relationship List $\{R\}$]

\textbf{Question}: [Question $Q$]

\textbf{Answer}: 

\end{mybox3}

\begin{figure*}
    \centering
    \includegraphics[width=1\linewidth]{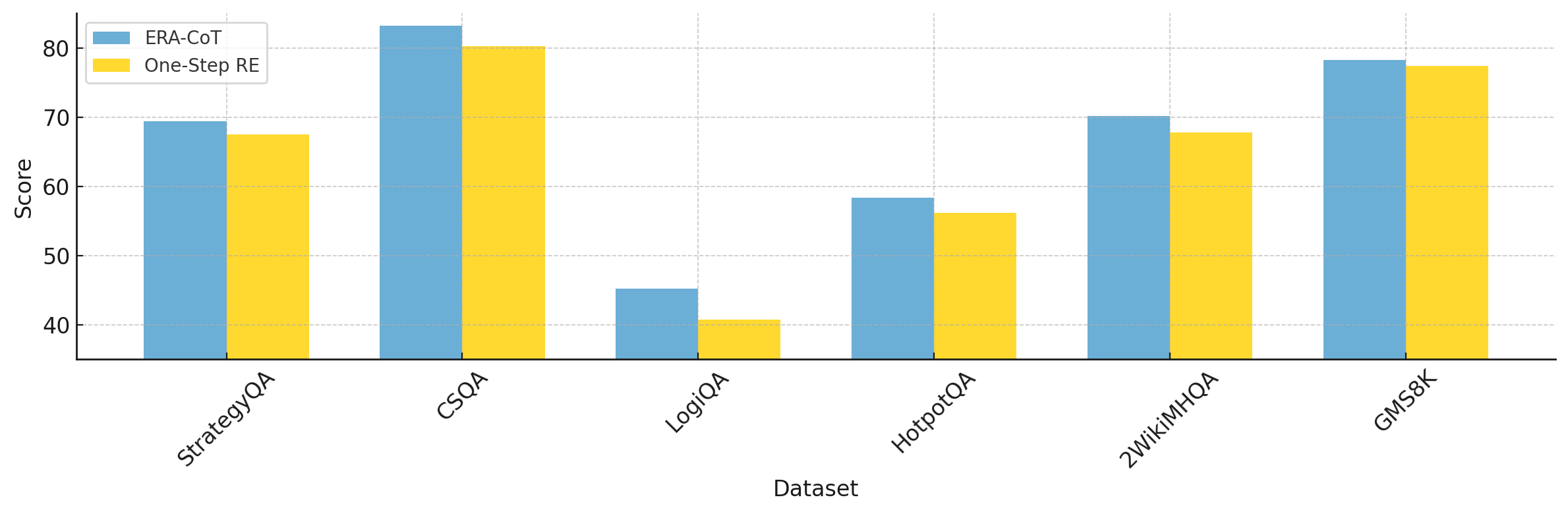}
    \caption{Comparison of performance between Single-Step and Multi-Step Relation Extraction. The score is evaluated on accuracy.}
    \label{fig:one-step-experiment}
\end{figure*}

\section{More Metrics on ERA-CoT}

To facilitate comparison with previous metrics and help understand how our method improves upon these baseline methods, we choose accuracy or Exact Match (EM) as the primary evaluation metric. To make the evaluation more persuasive, we compare different metrics and assess the \textbf{macro} scores across datasets including StrategyQA, LogiQA, and CommonsenseQA. For questions that do not include options, we use the standard F1 score. We present the results of GPT-3.5 in terms of precision, recall, and F1 score as Table~\ref{tab:f1}.

The outstanding performance demonstrated across various metrics in the evaluation highlights the effectiveness of our approach. This contributes to enhancing the model's ability to answer questions after extracting entities and relationships.

\begin{table}[]
\begin{tabular}{
>{\columncolor[HTML]{FFFDFA}}l 
>{\columncolor[HTML]{FFFDFA}}l 
>{\columncolor[HTML]{FFFDFA}}l 
>{\columncolor[HTML]{FFFDFA}}l }
\toprule
{\color[HTML]{333333} \textbf{Dataset}} & {\color[HTML]{333333} \textbf{F1}} & {\color[HTML]{333333} \textbf{Precision}} & {\color[HTML]{333333} \textbf{Recall}} \\ \midrule
{\color[HTML]{333333} StrategyQA}       & {\color[HTML]{333333} 70.2}        & {\color[HTML]{333333} 72.3}               & {\color[HTML]{333333} 69.6}            \\
{\color[HTML]{333333} CommonsenseQA}    & {\color[HTML]{333333} 82.1}        & {\color[HTML]{333333} 84.5}               & {\color[HTML]{333333} 80.2}            \\
{\color[HTML]{333333} HotpotQA}         & {\color[HTML]{333333} 74.6}        & {\color[HTML]{333333} 75.3}               & {\color[HTML]{333333} 80.5}            \\
{\color[HTML]{333333} LogiQA}           & {\color[HTML]{333333} 44.8}        & {\color[HTML]{333333} 43.7}               & {\color[HTML]{333333} 45.1}            \\
{\color[HTML]{333333} 2WikiMHQA}        & {\color[HTML]{333333} 81.5}        & {\color[HTML]{333333} 79.8}               & {\color[HTML]{333333} 86.5}            \\ \bottomrule
\end{tabular}
 \caption{
The evaluation of ERA-CoT on different datasets in terms of F1, precision, and recall.}
\label{tab:f1}
\end{table}

\section{Criterion on Discrimination Threshold}
\label{sec:criterion}

Regarding the relationship determination step, we establish the criteria for the relationship by providing corresponding score indicators. The specific indicator information is as follows:

\vspace{0.2cm}

\begin{itemize}
\item \texttt{Score 10: } \texttt{The implicit relationship between entities is very evident, almost attainable through one or two explicit relationship chains.}
\item \texttt{Score 8: } \texttt{There is a high likelihood that an implicit relationship exists between entities, deducible through a chain of explicit relationships.}
\item \texttt{Score 6: } \texttt{There is a probability that an implicit relationship exists between entities, with corresponding indications implied by context.}
\item \texttt{Score 4: } \texttt{The implicit relationship between entities may be correlated but cannot be defined.}
\item \texttt{Score 2: } \texttt{The implicit relationship between entities is largely unreliable.}
\item \texttt{Score 0: } \texttt{The implicit relationship between entities is completely unreliable.}
\end{itemize}

The information provided by such metrics is not fixed; we can preset other indicators or utilize ICL\citep{min-etal-2022-metaicl} to assist the model in scoring. In our experiment, we utilize the score 6 as based on the above criteria. Although the scoring effectiveness may not be optimal in this approach, by employing such benchmarks, we can ensure the elimination of some obvious errors or irrelevant contextual relationships. The results in Table~\ref{tab:Implicit Relationships} also demonstrate the necessity of this step (particularly when the inference of implicit relationship quantities increases, incorrect relationships can effectively be eliminated rather than forcing the model to generate irrelevant erroneous implicit relationships that would affect the final outcome due to the presence of relation inference).

\vspace{0.2cm}

\section{Why Not One-Step Relation Extraction?}

In the experiment of relation extraction, we complete it in three steps. The first step is explicit relation extraction. The second step is implicit relation inference based on the results of explicit relation extraction. The third step is to score the implicit relationships and remove the low-scoring situations that might lead to errors. Explicit relation extraction is defined as the corresponding relationships that LLMs or fine-tuned proprietary models can directly obtain from the context, and these relations must involve the context. 

For one-step relation extraction, GPT-RE \citep{wan2023gpt} provides a solution by extracting examples of relation extraction from questions that are similar to the inquiry, and then using few-shot prompting to assist LLMs in generating results for relation extraction. However, in the zero-shot situation, Vanilla RE instructs LLMs to search for all the relations of entities only based on the context and the entity list. This type of query is likely to cause errors, leading to a negative impact on answering questions.

\begin{mybox3}{\textbf{Entity Relation Extraction(One Step)}}

Given a sentence, and all entities within the sentence. Extract all relationships between entities in the sentence.

Every relationship stated as a triple: ($E_A$, $E_B$, $Relation$)

\textbf{Sentence}: [Sentence $S$]

\textbf{Entities}: [Entities List $\{E_i\}$]

\textbf{Relationships}:

\end{mybox3}

To better assess the necessity of dividing relation extraction into three steps, we conduct ablation experiments comparing situations with only explicit relation extraction, only implicit relation reasoning, and our method. The results highlight the importance of performing explicit relation extraction and implicit relation reasoning separately. Here, we will further compare single-step relation extraction with three-step relation extraction.

Figure~\ref{fig:one-step-experiment} shows the performance gap between single-step and multi-step extraction. In logical reasoning questions like those in LogiQA and HotpotQA, there is a significant performance gap, illustrating that multi-step relation extraction helps in accurately extracting relations and preventing erroneous information from affecting the model's answer, especially when complex relationships are involved. Especially in large language models, irrelevant information can easily influence the outcome \citep{shi2023large}, and incorrect relations are more likely to lead to erroneous answers.

\section{Example}

\begin{table*}[]
\resizebox{\linewidth}{!}{
\begin{tabular}{@{}p{20cm}@{}}
\toprule

\textbf{Context}:A Creative Commons license allows for the free distribution of an otherwise copyrighted piece of work, Works that are deemed in the public domain do not require a copyright or permission to use, Any work created before 1923 is in the public domain, Sandro Boticelli's The Birth of Venus painting was from 1486.\\
\textbf{Question}: Was a person sold a Creative Commons License for Boticelli's The Birth of Venus ripped off? \\
\textbf{Answer}: False. \\  \\ 

\textbf{Entities Extraction:}

Given a sentence, possible entities may include:[\textit{individuals, organizations, locations, ..., percentages}]. Find all entities based on the provided sentence.

\textbf{Sentence}: A Creative Commons license allows for...

\textbf{Entities}: [Creative Commons license, copyrighted piece of work,public domain, copyright, permission,work, 1923, public domain,Sandro Boticelli, The Birth of Venus painting, 1486]
\\ \\
\textbf{Explicit Relation Extraction:}

Given a sentence, and all entities within the sentence. Extract all relationships between entities which directly stated in the sentence.

Every relationship stated as a triple: ($E_A$, $E_B$, $Relation$)

\textbf{Sentence}: A Creative Commons license allows for...

\textbf{Entities}: [Creative Commons license, copyrighted piece of work, ...

\textbf{Relationships}:
(Creative Commons license, copyrighted piece of work, allows for)
(copyrighted piece of work, permission, require)
(Works, public domain, deemed)
(work, created before, 1923)
(work, public domain, is in)
(Sandro Boticelli's The Birth of Venus painting, was from, 1486)
\\ \\
\textbf{Implicit Relation Inference:}
Given a sentence, all entities, and all explicit relationships within the sentence. Infer all possible implicit relationships between entities. For each pair of entities, infer up to 3 implicit relationships.

Every relationship stated as a triple: ($E_A$, $E_B$, $Relation$)

\textbf{Sentence}: A Creative Commons license allows for ...

\textbf{Entities}: [Creative Commons license, ...

\textbf{Explicit Relationships}:(Creative Commons license, copyrighted piece of work, allows for)
(copyrighted piece of work, permission, require)...

\textbf{Implicit Relationships}:(Creative Commons license, copyrighted piece of work, is)
(Creative Commons license, permission, allows for)
(copyrighted piece of work, public domain, may be)
(public domain, require, not)
(copyrighted piece of work, created before, may be)
(public domain, created before, is)
(work, copyrighted piece of work, is)
(work, permission, requires)
(The Birth of Venus painting, copyright, may have)
(The Birth of Venus painting, permission, may require)
(Sandro Boticelli, The Birth of Venus painting, created)
(Sandro Boticelli, The Birth of Venus painting, is in)
(Sandro Boticelli, The Birth of Venus painting, is copyrighted)
\\ \\
\textbf{Relationship Discrimination:}

Given a sentence, and all uncertain relationships within the sentence. Score the confidence level of each relationship.

The confidence score ranges from 0 to 10, where a higher score indicates a higher likelihood of the relationship being correct.

Every relationship stated as a triple: ($E_A$, $E_B$, $Relation$)

\textbf{Sentence}: A Creative Commons license allows for ...

\textbf{Entities}: [Creative Commons license, ...

\textbf{Uncertain Relationships}:(Creative Commons license, copyrighted piece of work, is)
(Creative Commons license, permission, allows for)
(copyrighted piece of work, public domain, may be)...

\textbf{Scores}:[8,5,7,6,9,8,7,6,5,9,8,4]
\\ \\
\textbf{Question Answering:}

Given a sentence, all entities and all relationships within the sentence. Answering the question.

Every relationship stated as a triple: ($E_A$, $E_B$, $Relation$)

\textbf{Sentence}: A Creative Commons license allows for ...

\textbf{Entities}: [Creative Commons license, ...

\textbf{Relationships}:(Creative Commons license, copyrighted piece of work, allows for)
(copyrighted piece of work, permission, require)...

\textbf{Question}: Was a person sold a Creative Commons License for Boticelli's The Birth of Venus ripped off? 

\textbf{Answer}: 
No, a person who obtained a Creative Commons License for Boticelli's The Birth of Venus was not ripped off, because a Creative Commons license allows for the free distribution of copyrighted works, and Sandro Boticelli's The Birth of Venus painting was created in 1486, which is well before 1923, thus making it in the public domain and not requiring copyright or permission to use.

 \\ \bottomrule
\end{tabular}}
\caption{Case on StrategyQA.}
\label{tab:my-table}
\end{table*}

\end{document}